\documentclass[letterpaper]{article} 
\usepackage{aaai2026}  
\usepackage{times}  
\usepackage{helvet}  
\usepackage{courier}  
\usepackage[hyphens]{url}  
\usepackage{graphicx} 
\usepackage{natbib}  
\usepackage{caption} 
\usepackage{algorithm}
\usepackage{algorithmic}

\usepackage{newfloat}
\usepackage{listings}
\DeclareCaptionStyle{ruled}{labelfont=normalfont,labelsep=colon,strut=off} 
\floatstyle{ruled}
\newfloat{listing}{tb}{lst}{}
\floatname{listing}{Listing}
\usepackage{booktabs}
\usepackage{tabularx}
\usepackage{tabulary}
\usepackage{multirow}
\usepackage{amsbsy}
\usepackage{amssymb}
\usepackage{amsmath}
\usepackage{bm}
\usepackage{xcolor}
\usepackage{overpic}
\usepackage{bbding}

\nocopyright
\title{MDIQA: Unified Image Quality Assessment for\\Multi-dimensional Evaluation and Restoration}
\author{
    Shunyu Yao\textsuperscript{\rm 1},
    Ming Liu\textsuperscript{\rm 1},
    Zhilu Zhang\textsuperscript{\rm 1},
    Zhaolin Wan\textsuperscript{\rm 1},
    Zhilong Ji\textsuperscript{\rm 2},
    Jinfeng Bai\textsuperscript{\rm 2},
    Wangmeng Zuo\textsuperscript{\rm 1}
}
\affiliations{
    \textsuperscript{\rm 1}Harbin Institute of Technology\\
    \textsuperscript{\rm 2}Tomorrow Advancing Life\\
}

\begin{document}

\maketitle

\begin{abstract}
Recent advancements in image quality assessment (IQA), driven by sophisticated deep neural network designs, have significantly improved the ability to approach human perceptions.
However, most existing methods are obsessed with fitting the overall score, neglecting the fact that humans typically evaluate image quality from different dimensions before arriving at an overall quality assessment.
To overcome this problem, we propose a multi-dimensional image quality assessment (MDIQA) framework.
Specifically, we model image quality across various perceptual dimensions, including five technical and four aesthetic dimensions, to capture the multifaceted nature of human visual perception within distinct branches.
Each branch of our MDIQA is initially trained under the guidance of a separate dimension, and the respective features are then amalgamated to generate the final IQA score.
%
Additionally, when the MDIQA model is ready, we can deploy it for a flexible training of image restoration (IR) models, enabling the restoration results to better align with varying user preferences through the adjustment of perceptual dimension weights.
Extensive experiments demonstrate that our MDIQA achieves superior performance and can be effectively and flexibly applied to image restoration tasks.
The code is available: https://github.com/YaoShunyu19/MDIQA.
\end{abstract}


\section{Introduction}
Image Quality Assessment (IQA) is a pivotal task in computer vision, focused on automatically evaluating the perceptual quality of images in alignment with subjective human perception.
Its wide-ranging significance spans numerous applications in image processing, offering critical insights into the effectiveness of image enhancement and restoration algorithms.
With the advancements in image processing technologies, IQA methods have evolved considerably over the past few decades, leading to more accurate and reliable quality assessments.

\begin{figure}[t]
\scriptsize{
\begin{center}

\begin{overpic}[width=0.47\textwidth]{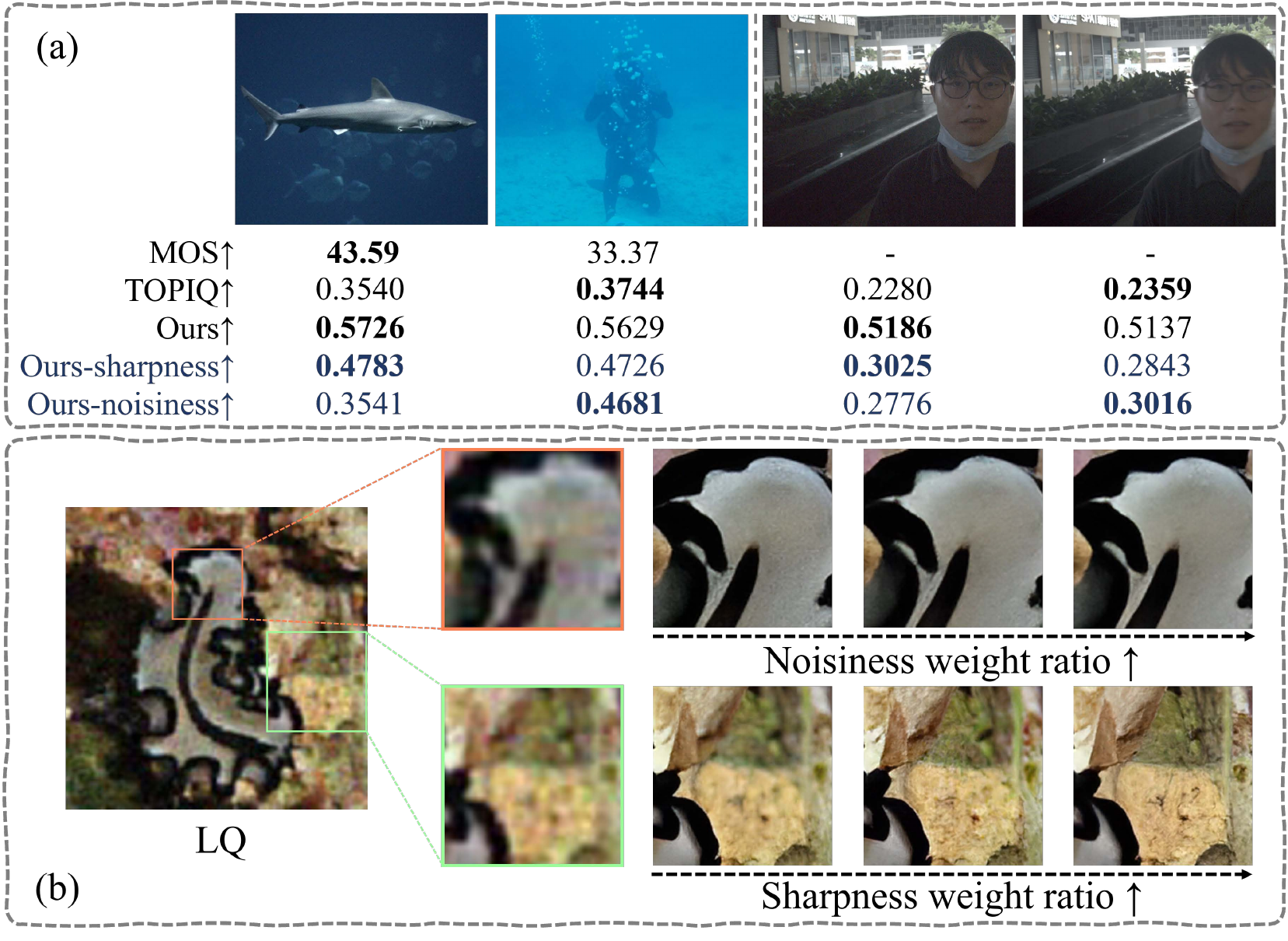}
\end{overpic}
\caption{
(a) For Evaluation: our MDIQA achieves higher consistency with human subjective opinions and provides interpretable predictions through multi-dimensional quality modeling, in contrast to existing methods that output a single, opaque score.
(b) For Restoration: our MDIQA serves as a tunable loss for image restoration, where adjusting the weights of perceptual dimensions (e.g., sharpness or noisiness) enables adaptation to varying restoration preferences.
Please zoom in for better view.
}
\label{fig:intro}
\end{center}}
\end{figure}


Existing IQA methods, including those based on traditional statistical metrics \cite{SSIM,MS-SSIM,VIF,IFC} and increasingly sophisticated deep learning approaches \cite{DB-CNN,DISTS,MUSIQ,MANIQA}, have significantly improved the alignment between quality assessments and human subjective opinions, thereby facilitating progress in various image processing tasks. However, most of these methods share a fundamental limitation: they compress rich information into a single overall score.
In contrast, human assessment of image quality, guided by the perceptual mechanisms of the human visual system (HVS) \cite{Foundations-of-Vision}, naturally integrates multiple dimensions, such as sharpness, noise, etc.
Recently, several efforts have begun to explore multi-dimensional quality evaluation.
SPAQ \cite{SPAQ} introduced an IQA dataset with multi-attribute ratings and EXIF tags, enabling the study of interactions between perceptual image quality and factors such as camera settings and scene categories. PARA \cite{PARA}, focusing on image aesthetic assessment (IAA), provided rich attribute annotations to investigate the interplay between personalized aesthetic preferences and human subjective attributes.
While these works have shed light on the relationships between perceptual dimensions and contextual factors, they primarily remain at the level of dataset construction and empirical analysis. Specifically, they do not fully leverage multi-dimensional annotations to develop more accurate or interpretable models for quality prediction.
In the video quality domain, MaxVQA \cite{MaxVQA} assesses quality across similar perceptual aspects, but tends to overlook content semantics due to fragment-based sampling, and lacks explicit modeling of the relationship between sub-dimensions and overall quality.
Overall, the potential of multi-dimensional quality modeling remains underexplored, particularly in improving assessment performance through effective integration of multiple dimensions, and in enabling practical downstream applications.

In this paper, we propose a multi-dimensional image quality assessment (MDIQA) framework that aligns with the human tendency to assess image quality across diverse perceptual dimensions.
Our approach organizes the evaluation process into two primary categories: technical and aesthetic following DOVER \cite{DOVER}, and further refines them into five technical dimensions and four aesthetic dimensions to capture fine-grained aspects of human perception.
To model these dimensions, we design a multi-branch network in which the technical and aesthetic components have separate backbones, each assigned across multiple dimension-specific heads. These components are pretrained on richly annotated datasets to enable robust representation learning for both technical fidelity and aesthetic quality.
To produce an overall quality score in a meaningful and coherent way, we introduce a dynamic weighting mechanism that generates an image-specific weight vector, allowing the model to adaptively combine the dimension-wise predictions based on content characteristics. We further enhance semantic awareness by injecting CLIP \cite{CLIP} visual features into each head.
By integrating dimension-aware modeling, adaptive weighting, and semantic enrichment, our MDIQA framework addresses key limitations of existing single-score IQA methods. It offers both fine-grained quality analysis and improved accuracy in overall score prediction, demonstrating the dual value of multi-dimensional image quality assessment.

Furthermore, our MDIQA framework offers a versatile loss function for Image Restoration (IR) tasks.
Unlike most existing methods such as PSNR and LPIPS \cite{LPIPS}, which rely on fixed statistical or perceptual differences between the restored image and a pristine-quality reference, our approach enables dynamic adjustment of the weights assigned to each perceptual dimension.
This flexibility effectively addresses the limitations of traditional loss functions by allowing the optimization process to focus on more relevant aspects of perceptual quality.
Specifically, the contribution of each dimension can be manually controlled by scaling its corresponding weight during training. For example, increasing the sharpness weight guides the model to prioritize edge preservation over noise reduction.
This mechanism enables the model to prioritize dimensions that better reflect specific restoration goals or user preferences, facilitating more targeted optimization.
Such controllability ensures that the restoration results are better aligned with desired quality attributes, offering a more adaptive and interpretable alternative to conventional loss designs. By dynamically adjusting the dimension weights, our method tailors the restoration process to diverse perceptual needs.

Extensive experiments validate the effectiveness of our method from two key perspectives.
Our MDIQA achieves state-of-the-art performance on several widely used IQA datasets, including CLIVE \cite{CLIVE}, KonIQ-10k \cite{KonIQ-10k}, SPAQ \cite{SPAQ}, and FLIVE \cite{PaQ2PiQ}, highlighting the benefits of the multi-dimensional evaluation paradigm and the strength of our network design. Fig. \ref{fig:intro}(a) qualitatively illustrates the advantages of MDIQA in assessment accuracy and interpretability.
Furthermore, both quantitative results and qualitative comparisons demonstrate that, when applied to image restoration tasks, the adjustable dimension weights in our framework enable flexible control over restoration, leading to outputs better aligned with diverse user preferences and task-specific quality requirements. Fig. \ref{fig:intro}(b) clearly demonstrates how the tunable loss effectively guides restoration.

Our main contributions can be summarized as follows:
\begin {itemize}
   \item We propose MDIQA, a multi-dimensional image quality assessment framework that leverages diverse perceptual dimensions to support fine-grained quality analysis and facilitate more accurate overall quality prediction.
   \item Our MDIQA can be deployed as an adjustable loss function for IR, which dynamically adapts dimension weights based on specific requirements, offering a novel mechanism to align restoration results with targeted needs.
   \item Extensive experiments demonstrate that our MDIQA achieves state-of-the-art performance on several widely used IQA datasets and can be effectively applied to IR, adjusting to varying user requirements.
\end {itemize}

\section{Related Work}
\subsection{Image Quality Assessment Methods}
Over the past few decades, IQA algorithms have seen remarkable progress.
Traditional methods, such as PSNR and SSIM variants \cite{SSIM,MS-SSIM,CW-SSIM}, quantify quality by measuring statistical differences between distorted images and pristine-quality references, while NIQE \cite{NIQE} evaluates quality by leveraging the Natural Scene Statistics (NSS), comparing the distribution of a given image with natural images.
With the rise of deep learning, IQA methods have predominantly adopted end-to-end frameworks, ranging from CNN-based to Transformer-based models, further pushing the boundaries of IQA performance.
Early CNN-based methods \cite{DeepIQA,DISTS,WADIQAM} utilized pre-trained networks \cite{AlexNet,VGG} to map deep image features to perceptual scores. Subsequently, Transformer-based models \cite{MUSIQ,MANIQA} improved performance by capturing global dependencies. More recently, hybrid architectures that integrate the strengths of both CNNs and Transformers, including AHIQ \cite{AHIQ} and LoDa \cite{LoDa}, have advanced the field and achieved state-of-the-art performance.

However, these approaches typically generate a single overall quality score, neglecting the multi-dimensional nature of human perception in image quality assessment.
Recent efforts \cite{SPAQ} have explored multi-dimensional quality labeling, analyzing the interplay between perceptual dimensions to improve interpretability and examining their relationships with factors such as EXIF tags.
Some studies have further extended this approach to video quality assessment \cite{MaxVQA} and image aesthetic assessment \cite{PARA,Aesexpert}.
Despite these advancements, most existing multi-dimensional quality assessment studies have primarily focused on exploring the relationships between perceptual dimensions and external factors. However, they have not fully utilized multi-dimensional annotations to enhance the overall accuracy of quality assessment. In this work, we bridge this gap by introducing a multi-dimensional IQA network that achieves both enhanced performance and interpretable quality assessment.

\subsection{Loss Functions in Image Restoration}

In image restoration, the design of loss functions plays a pivotal role in determining the perceptual quality of the restored images.
Early approaches typically relied on pixel-level losses such as mean squared error (MSE), which often led to overly smooth results, especially in high-frequency regions.
To produce more realistic restorations, the introduction of generative adversarial networks (GANs) \cite{GAN} marked a significant shift in loss function design.
SRGAN \cite{SRGAN} demonstrated that GAN-based losses could guide image reconstructions toward photo-realistic solutions by encouraging alignment with the natural image manifold.
Subsequent super-resolution methods \cite{ESRGAN,RankSRGAN} further embraced GAN losses to enhance the realism of restored outputs.
In parallel, perceptual losses \cite{Perceptual-losses} and LPIPS \cite{LPIPS}, which measure Euclidean distances between features extracted from pre-trained deep networks \cite{AlexNet,VGG}, have proven effective in capturing high-level perceptual and semantic differences.
Consequently, perceptual loss has become a standard component in many image restoration frameworks \cite{Real-ESRGAN,HGGT,zhang2024self}.
In addition, to enforce fidelity in specific perceptual dimensions, Ignatov et al. \cite{color_loss} proposed a color loss that applies Gaussian blur to both the restored and ground-truth images before computing the L2 distance, thereby ensuring consistency from a color perspective.
Despite their effectiveness, these approaches are generally built upon fixed evaluation criteria, lacking the flexibility to accommodate varying user preferences and task-specific requirements.

\section{Proposed Method}

\subsection{Preliminaries}
\begin{figure*}
\scriptsize{
\begin{center}

\begin{overpic}[width=0.97\textwidth]{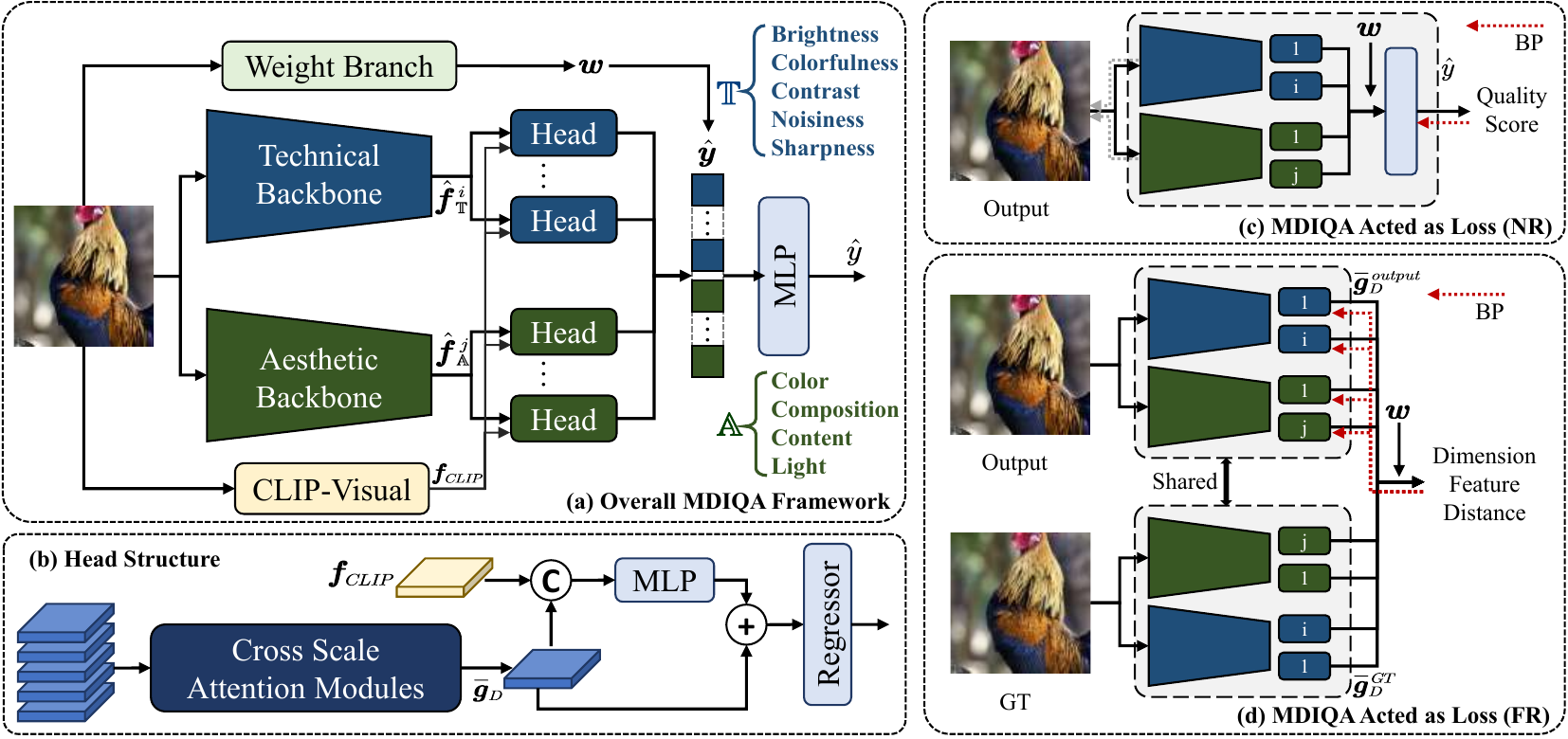}
\end{overpic}

\caption{Illustration of our MDIQA. (a) Overall architecture of MDIQA. (b) The structure of each head. (c) MDIQA acted as No-Reference loss function for image restoration. (d) MDIQA acted as Full-Reference loss function for image restoration.}
\label{fig:method}

\end{center}}
\end{figure*}

In real-world IQA scenarios, reference images are rarely available for assessing degraded content. We therefore focus on the No-Reference (NR) setting, where perceptual quality must be predicted directly from the degraded image alone.
To better align with human perception, we decompose image quality into five technical dimensions and four aesthetic dimensions, denoted as $\mathbb{T}$ and $\mathbb{A}$, respectively, as illustrated in Fig.~\ref{fig:method}(a).
Our MDIQA jointly models these aspects to produce a comprehensive quality score.
We further extend this multi-dimensional representation to Image Restoration (IR), where it serves as a tunable loss. By adjusting the weights of individual dimensions, the restoration process can be adapted to diverse perceptual preferences.


\subsection{Multi-Dimensional Image Quality Assessment}
\label{architecture}

\subsubsection{Overall Architecture}

As illustrated in Fig. \ref{fig:method}(a) and (b), MDIQA employs two shared backbones to extract multi-level features for technical and aesthetic dimensions separately, considering their distinct perceptual characteristics while maintaining computational efficiency.
To ensure dimension-specific modeling, lightweight heads process the backbone features, with CLIP visual features incorporated into each head to enrich semantic context understanding.
Simultaneously, an efficient image encoder predicts a weight vector that adapts to the input image, reflecting the varying importance of each dimension.
These weights are used to refine the individual dimension scores, which are then fused through a MLP to produce the final overall quality score.

\subsubsection{Shared Backbone Structure}

To efficiently extract deep image features while maintaining computational feasibility, we adopt a dual-backbone structure, as shown in Fig. \ref{fig:method}(a). While assigning an independent network to each dimension would be conceptually straightforward, it becomes computationally prohibitive as the number of dimensions increases.
Instead, we employ two shared backbones: one dedicated to technical dimensions and the other to aesthetic dimensions. This design ensures that features are well-aligned with their respective quality aspects while significantly reducing model complexity.
Specifically, following common practice in perceptual modeling \cite{LPIPS}, we use pre-trained CNNs to extract multi-scale features from input images. These features capture rich local distortion patterns, which are essential for accurate quality prediction.
To unify the multi-scale representations into a consistent spatial format, we apply Gated Local Pooling (GLP) modules \cite{TOPIQ} at each scale. These modules enhance quality-aware information while suppressing redundant features, resulting in more compact and informative feature maps.

\subsubsection{Specific Head Structure}
Each quality dimension requires a specialized representation that captures its unique perceptual attributes. To achieve this, we design a set of lightweight and independent heads, with each head dedicated to a specific technical or aesthetic dimension. This structure allows the model to effectively capture the distinct characteristics of each quality aspect.
In our framework, the multi-scale features extracted by the shared backbones are fed into their corresponding heads. Within each head, we apply Cross-Scale Attention Modules (CSAM) \cite{TOPIQ}, which fuse features across different levels to generate more informative and discriminative representations, as illustrated in Fig. \ref{fig:method}(b).
To further enrich the representations with semantic context, we incorporate CLIP visual features into each head, which are concatenated with the fused feature maps and refined through an MLP with a residual connection to improve high-level content understanding.
Finally, each head concludes with a dedicated regressor composed of global pooling and MLP, which maps the processed features to a quality score for that dimension.

\subsubsection{Weighted Quality Regression}

Given the predicted dimension scores, a direct solution is to fuse them into an overall quality score using a simple MLP. However, the relative importance of each dimension can vary significantly depending on the image content and visual characteristics.
To address this, we introduce a lightweight weighting branch that generates an image-adaptive weight vector $\bm{w}$ , which modulates the dimension scores before fusion. The weighted scores are then passed through a 3-layer MLP to produce the final quality prediction.

\subsubsection{Loss Functions}
Building on the insights from \cite{TOPIQ, MIPM}, our MDIQA adopts a unified hybrid loss that combines Mean Squared Error (MSE) with the Norm-in-Norm (NiN) loss \cite{NiNLoss}, as detailed in the supplementary materials.


\subsection{Tunable Loss for Image Restoration}
\label{loss4IR}
As illustrated in Fig. \ref{fig:method}(c) and (d), our MDIQA can serve as a tunable loss for image restoration tasks.
To demonstrate its effectiveness, we integrate MDIQA into Real-ESRGAN \cite{Real-ESRGAN}, a widely used image restoration model.
The original loss function of Real-ESRGAN, denoted as $\mathcal{L}_{org}$, combines L1 loss, perceptual loss, and GAN loss.
We extend this by introducing two variants of the MDIQA-based loss: a No-Reference (NR) version and a Full-Reference (FR) version.
In the NR variant, we compute the quality score of the restored image $R$ without relying on ground truth (GT) $H$. The corresponding loss is defined as:
\begin{equation}
\label{eq:nr_loss}
\mathcal{L}_{NR}=-\tfrac{1}{B}{\sum}_{i=1}^B f(R_i,\bm{w}), 
\end{equation}
where $f(R_i,\bm{w})$ denotes the quality score of $R_i$, with higher values indicating better perceptual quality. This formulation enables the model to optimize restoration based solely on the learned quality assessment.

For scenarios where reference images are available, we further design an FR loss that minimizes the feature distance between the restored image $R$ and the ground truth $H$ across all perceptual dimensions. Specifically, we compute the L1 distance between dimension-specific features and weight them using the dimension weights $\bm{w}$. The FR loss is formulated as:
\begin{equation}
\label{eq:fr_loss}
\mathcal{L}_{FR}=\tfrac{1}{B}{\sum}_{i=1}^B {\sum}_{D\in \mathbb{T}\cup \mathbb{A}} \bm{w}_D \mathcal{L}_1(\bar{\bm{g}}_D^{R},\bar{\bm{g}}_D^{H}),
\end{equation}
Where $\bar{\bm{g}}_D^{R}$ and $\bar{\bm{g}}_D^{H}$ denote the dimension-specific features of the restored and the ground truth, respectively.

A key advantage of our framework lies in its tunable dimension weighting mechanism, which allows for manual adjustment of the weight vector $\bm{w}$ to emphasize specific perceptual dimensions during training.
For instance, increasing the sharpness weight as $\bm{w}'_{sharpness}=\lambda 
\bm{w}_{sharpness}$ with $\lambda >1$ enhances the model's focus on preserving fine textures.
Conversely, boosting the weights associated with noise results in smoother output images. This flexible control enables the restoration process to be tailored to application-specific requirements, whether the priority is detail preservation, noise suppression, or balanced enhancement. Such adaptability underscores the versatility and practical value of our MDIQA framework.

\section{Experiments}
\label{sec:experiments}

\subsection{Experiment Settings}
\subsubsection{Datasets}
We evaluate our method using four publicly available authentic IQA datasets, ie., CLIVE \cite{CLIVE}, KonIQ-10k \cite{KonIQ-10k}, SPAQ \cite{SPAQ}, and FLIVE \cite{PaQ2PiQ}.
While FLIVE provides an official train-test split, for the other three datasets, we randomly partition the data into training and testing sets at an 8:2 ratio, repeating this process 10 times to reduce potential bias, with final results reported as the average across all runs.
Further details regarding the datasets can be found in the supplementary materials.

\subsubsection{Evaluation Criteria}
To evaluate the performance of IQA algorithms, we employ two widely used metrics: Spearman's Rank Correlation Coefficient (SRCC) for measuring prediction monotonicity and Pearson's Linear Correlation Coefficient (PLCC) for quantifying prediction accuracy.

\subsubsection{Implementation Details}

\begin{table}[t]
    \centering
    \setlength{\tabcolsep}{2pt}
    \caption{Method comparisons on CLIVE, KonIQ-10k, SPAQ and FLIVE. The best results are highlighted in bold, and the second-best results are underlined.}
    \label{tab:biqa_benchmarks}
    
    \scalebox{0.82}{
    \begin{tabulary}{\textwidth}{@{}Lcccc@{}}
        \toprule
        \multirow{2}{*}{Method} & CLIVE & KonIQ-10k & SPAQ & FLIVE \\
        & \multicolumn{1}{c}{\footnotesize SRCC$\uparrow$/PLCC$\uparrow$} & \multicolumn{1}{c}{\footnotesize SRCC$\uparrow$/PLCC$\uparrow$} & \multicolumn{1}{c}{\footnotesize SRCC$\uparrow$/PLCC$\uparrow$} & \multicolumn{1}{c}{\footnotesize SRCC$\uparrow$/PLCC$\uparrow$} \\
        \midrule
        DIIVINE & 0.588/0.591 & 0.546/0.558 & 0.599/0.600 & 0.092/0.186 \\
        BRISQUE & 0.629/0.629 & 0.681/0.685 & 0.809/0.817 & 0.303/0.341 \\
        NIQE & 0.451/0.493 & 0.377/0.389 & 0.697/0.685 & 0.288/0.211 \\
        ILNIQE & 0.508/0.508 & 0.523/0.537 & 0.713/0.721 & 0.294/0.332 \\
        PI & 0.462/0.521 & 0.457/0.488 & 0.709/0.724 & 0.170/0.334 \\
        BMPRI & 0.487/0.523 & 0.658/0.655 & 0.750/0.754 & 0.274/0.315 \\
        \midrule
        CNNIQA & 0.627/0.601 & 0.685/0.684 & 0.796/0.799 & 0.306/0.285 \\
        MEON & 0.697/0.710 & 0.611/0.628 & -/- & 0.365/0.394 \\
        WaDIQaM & 0.682/0.671 & 0.804/0.807 & 0.840/0.845 & 0.455/0.467 \\
        SFA & 0.804/0.821 & 0.888/0.897 & 0.906/0.907 & 0.542/0.526 \\
        DBCNN & 0.869/0.869 & 0.875/0.884 & 0.911/0.915 & 0.545/0.551 \\
        HyperIQA & 0.859/0.882 & 0.906/0.917 & 0.916/0.919 & 0.535/0.623 \\
        MetaIQA & 0.835/0.802 & 0.887/0.856 & -/- & 0.540/0.507 \\
        TIQA & 0.845/0.861 & 0.892/0.903 & -/- & 0.541/0.581 \\
        TReS & 0.846/0.877 & 0.915/0.928 & -/- & 0.554/0.625 \\
        MUSIQ & -/- & 0.916/0.928 & 0.917/0.920 & \underline{0.646}/\underline{0.739} \\
        Re-IQA & 0.840/0.854 & 0.914/0.923 & 0.918/0.925 & 0.645/0.733 \\
        QPT & \textbf{0.895}/\textbf{0.914} & 0.927/0.941 & \underline{0.925}/\underline{0.928} & 0.610/0.677 \\
        TOPIQ & 0.870/0.884 & 0.926/0.939 & 0.921/0.924 & 0.633/0.722 \\
        QCN & 0.875/0.893 & \underline{0.934}/\underline{0.945} & 0.923/0.928 & 0.644/\textbf{0.741} \\
        LoDa & 0.876/\underline{0.899} & 0.932/0.944 & \underline{0.925}/\underline{0.928} & 0.631/0.726 \\
        Ours & \underline{0.890}/\underline{0.899} & \textbf{0.948}/\textbf{0.956} & \textbf{0.928}/\textbf{0.931} & \textbf{0.647}/0.734 \\
        \bottomrule
    \end{tabulary}}

\end{table}


Our MDIQA is implemented in two stages. In the first stage, the backbones and heads are trained on multi-dimensional annotated datasets SPAQ and PARA without CLIP feature injection, allowing each head to learn dimension-specific perceptual characteristics. In the second stage, we incorporate the weight branch and CLIP (ResNet50) feature injection into the model. The backbones and cross-scale attention modules are frozen, while we focus on training the weight branch and the multi-dimensional representation aggregation component.

During IQA training, we use the AdamW optimizer with a batch size of 16. Images are randomly cropped to $384\times 384$ and augmented with random horizontal flipping. In the first stage, the learning rate is set to $3e-5$ with a weight decay of $1e-5$, and the technical and aesthetic dimensions are trained for 8 and 4 epochs, respectively. In the second stage, the learning rate is reduced to $1e-5$, and training continues for 5 epochs. The CosineAnnealingLR scheduler is used to adjust the learning rate throughout both stages.

For the application in the image restoration task, we use Real-ESRGAN as the baseline. The Adam optimizer is employed with a learning rate of $3e-5$ and a batch size of 12. The model is fine-tuned on the DF2K-OST \cite{Real-ESRGAN} dataset for 60,000 iterations using pre-trained parameters.
The MDIQA model trained on the KonIQ-10k dataset serves as the loss function, with coefficients for $\mathcal{L}_{NR}$ and $\mathcal{L}_{FR}$ set to 1.0 and 5.0, respectively, while Real-ESRGAN's original loss ($\mathcal{L}_{org}$) retains its default settings.
%
All experiments are conducted on an NVIDIA V100 GPU.

\subsection{Comparison with SOTA IQA Methods}
\subsubsection{Evaluation on Authentic Benchmarks}
\label{sec:nr_benchmark}

We conduct a comprehensive evaluation of our proposed model by comparing it with 6 traditional algorithms (DIIVINE \cite{DIIVINE}, BRISQUE \cite{BRISQUE}, NIQE \cite{NIQE}, ILNIQE \cite{ILNIQE}, PI \cite{PI}, and BMPRI \cite{BMPRI}) and 15 deep learning-based methods (CNNIQA \cite{CNNIQA}, MEON \cite{MEON}, WaDIQaM \cite{WADIQAM}, SFA \cite{SFA}, DBCNN \cite{DB-CNN}, HyperIQA \cite{hyperIQA}, MetaIQA \cite{MetaIQA}, TIQA \cite{TIQA}, TReS \cite{TReS}, MUSIQ \cite{MUSIQ}, Re-IQA \cite{Re-IQA}, QPT \cite{QPT}, QCN \cite{QCN}, and LoDa \cite{LoDa}) across 4 authentic datasets, as shown in Table \ref{tab:biqa_benchmarks}.
Compared to traditional methods that rely on natural scene statistics (NSS), learning-based approaches provide quality predictions that better align with human perception, leading to superior performance across benchmark datasets. Among these, recent models such as MUSIQ, TOPIQ, and LoDa show progressively improved accuracy, demonstrating the benefits of more advanced network architectures.
Besides, we observe that pre-training strategies enhance performance.
Specifically, QPT achieves the highest SRCC and PLCC on CLIVE, owing to its self-supervised learning mechanism that synthesizes a wide variety of degraded images.
In contrast, other learning-based algorithms are more prone to overfitting on CLIVE.
However, our method achieves state-of-the-art results on KonIQ-10k, SPAQ, and the SRCC metric on FLIVE, and competitive results on CLIVE.
These outcomes not only highlight the strengths of our MDIQA but also provide valuable insights for the future advancement of IQA.

\subsubsection{Cross-Dataset Evaluation}

\begin{table}[t]
    \centering
    \small
    \setlength{\tabcolsep}{2pt}
    \caption{SRCC$\uparrow$ scores on the cross-dataset benchmark for performance evaluation.}
    \label{tab:cross-dataset}
    \renewcommand{\arraystretch}{1.0}
    
    \scalebox{0.82}{
    \begin{tabulary}{\textwidth}{@{}lcccccc@{}}
        \toprule
        Train on & \multicolumn{2}{c}{KonIQ-10k} & \multicolumn{2}{c}{FLIVE} & \multicolumn{2}{c}{SPAQ} \\
        \cmidrule(rl{0pt}){2-3}
        \cmidrule(rl{0pt}){4-5}
        \cmidrule(rl{0pt}){6-7}
        Test on & CLIVE & FLIVE & CLIVE & KonIQ-10k & KonIQ-10k & FLIVE \\
        \midrule
        WaDIQaM & 0.682 & - & 0.699 & 0.708 & - & - \\
        DBCNN & 0.755 & - & 0.724 & 0.716 & - & - \\
        P2P-BM & 0.770 & - & 0.738 & 0.755 & - & - \\
        HyperIQA & 0.785 & - & 0.735 & 0.758 & - & - \\
        TReS & 0.777 & 0.492 & 0.734 & 0.707 & - & - \\
        MUSIQ & 0.789 & 0.498 & 0.767 & 0.708 & 0.680 & 0.563 \\
        TOPIQ & \underline{0.821} & \underline{0.580} & 0.787 & 0.762 & \underline{0.763} & \underline{0.565} \\
        LoDa & 0.811 & - & \underline{0.805} & \underline{0.763} & - & \\
        Ours & \textbf{0.824} & \textbf{0.628} & \textbf{0.811} & \textbf{0.783} & \textbf{0.803} & \textbf{0.619} \\
        \bottomrule
    \end{tabulary}}

\end{table}

To evaluate the generalization capability of our method, we conduct cross-dataset experiments following the evaluation protocol established in TOPIQ, as presented in Table \ref{tab:cross-dataset}. Our MDIQA is compared with existing methods including WaDIQaM, DBCNN, P2P-BM \cite{PaQ2PiQ}, HyperIQA, TReS, MUSIQ, TOPIQ, and LoDa. The results demonstrate that our approach consistently achieves superior performance across diverse dataset configurations, significantly outperforming existing methods. This robust performance underscores the effectiveness of MDIQA in handling both varied distortion domains and substantial content variations.
Moreover, the superior performance across datasets with distinct quality characteristics indicates the practical potential of MDIQA in real-world applications.
The integration of the multi-dimensional evaluation paradigm and adaptive weighting mechanisms enables our method to better capture the complex interplay between diverse perceptual factors, making it particularly suitable for practical scenarios with diverse quality criteria.

\subsection{Ablation Studies}
\subsubsection{Ablation of the Network Structure}

Based on the results in Table \ref{tab:ablation_network_structure}, we conduct an ablation study to evaluate the impact of key components in our network structure, including the weight branch, regressor fine-tuning in the heads, and CLIP feature injection.
The first row presents the baseline results, where a simple MLP is used to fuse the scores from different dimensions.
Introducing the lightweight weight branch significantly improves performance, demonstrating the effectiveness of image-adaptive dimension weighting in multi-dimensional IQA.
Furthermore, replacing scalar fusion with feature-level fusion via head regressor training enhances the model’s representational capacity while preserving dimension decoupling, leading to further gains.
Finally, incorporating CLIP features improves spatial semantic understanding, yielding the best overall performance. In particular, when testing across datasets, CLIP brings 0.012 SRCC improvements on  FLIVE dataset (see Sec.\~D.3 in suppl.).
These results collectively validate the contributions of each component in enhancing the quality assessment process.

\begin{table}[!htbp]
    \small
    \centering
    \caption{Performance with ablation studies about network structure performed on KonIQ-10k and FLIVE.}
    \label{tab:ablation_network_structure}
    
    \scalebox{0.95}{
    \begin{tabulary}{\textwidth}{@{\hspace{2pt}}c@{\hspace{2pt}}c@{\hspace{2pt}}c@{\hspace{6pt}}c@{\hspace{2pt}}c@{\hspace{2pt}}c@{\hspace{2pt}}c@{\hspace{2pt}}}
        \toprule
        \multirow{2}{*}{\parbox{1cm}{\centering Weight\\Branch}} & \multirow{2}{*}{\parbox{1.8cm}{\centering Fine-tuning\\Regressor}} & \multirow{2}{*}{\parbox{1cm}{\centering CLIP\\Features}} & \multicolumn{2}{c}{KonIQ-10k} & \multicolumn{2}{c}{FLIVE} \\
        & & & SRCC$\uparrow$ & PLCC$\uparrow$ & SRCC$\uparrow$ & PLCC$\uparrow$ \\
        \midrule
        \centering \XSolidBrush & \centering \XSolidBrush & \centering \XSolidBrush & 0.894 & 0.921 & 0.602 & 0.686 \\
        \centering \Checkmark & \centering \XSolidBrush & \centering \XSolidBrush & 0.941 & 0.950 & 0.636 & 0.711 \\
        \centering \Checkmark & \centering \Checkmark & \centering \XSolidBrush & 0.946 & 0.955 & 0.645 & 0.732 \\
        \centering \Checkmark & \centering \Checkmark & \centering \Checkmark & \textbf{0.948} & \textbf{0.956} & \textbf{0.647} & \textbf{0.734} \\
        \bottomrule
    \end{tabulary}}

\end{table}

\subsubsection{Performances with Different Dimensions}

We evaluate the impact of different dimensions by comparing various combinations of technical and aesthetic dimensions in Table \ref{tab:ablation_dimensions}.
The first row uses all five technical dimensions without any aesthetic dimensions and already achieves strong performance. When only aesthetic dimensions are used, the results on KonIQ-10k remain comparable, while performance on FLIVE improves, likely due to the inclusion of images from the AVA dataset \cite{AVA}, which emphasizes aesthetic quality.
Combining all five technical and four aesthetic dimensions yields the best results, further demonstrating the effectiveness of our MDIQA framework in capturing both perceptual aspects for accurate image quality assessment.

\begin{table}[!htbp]
    \small
    \centering
    \caption{Performance with ablation studies about different dimensions performed on KonIQ-10k and FLIVE.}
    \label{tab:ablation_dimensions}
    
    \scalebox{0.95}{
    \begin{tabulary}{\textwidth}{@{\hspace{2pt}}c@{\hspace{6pt}}c@{\hspace{6pt}}c@{\hspace{2pt}}c@{\hspace{2pt}}c@{\hspace{2pt}}c@{\hspace{2pt}}}
        \toprule
        \multirow{2}{*}{Technical} & \multirow{2}{*}{Aesthetic} & \multicolumn{2}{c}{KonIQ-10k} & \multicolumn{2}{c}{FLIVE} \\
        & & SRCC$\uparrow$ & PLCC$\uparrow$ & SRCC$\uparrow$ & PLCC$\uparrow$ \\
        \midrule
        \centering \Checkmark & \centering \XSolidBrush & 0.939 & 0.947 & 0.634 & 0.727 \\
        \centering \XSolidBrush & \centering \Checkmark & 0.937 & 0.949 & 0.643 & 0.731 \\
        \centering \Checkmark & \centering \Checkmark & \textbf{0.948} & \textbf{0.956} & \textbf{0.647} & \textbf{0.734} \\
        \bottomrule
    \end{tabulary}}

\end{table}

\subsection{Results in Image Restoration Applications}

To evaluate the effectiveness of our MDIQA loss in image restoration, we use widely adopted NR-IQA metrics such as NIQE, CLIP-IQA \cite{CLIPIQA}, and MUSIQ, along with dimension-specific and overall quality scores predicted by MDIQA. We also consider qualitative visual results to assess perceptual improvements. More results are provided in the supplementary materials.

\begin{figure}[t]
\scriptsize{
\begin{center}

\begin{overpic}[width=0.45\textwidth]{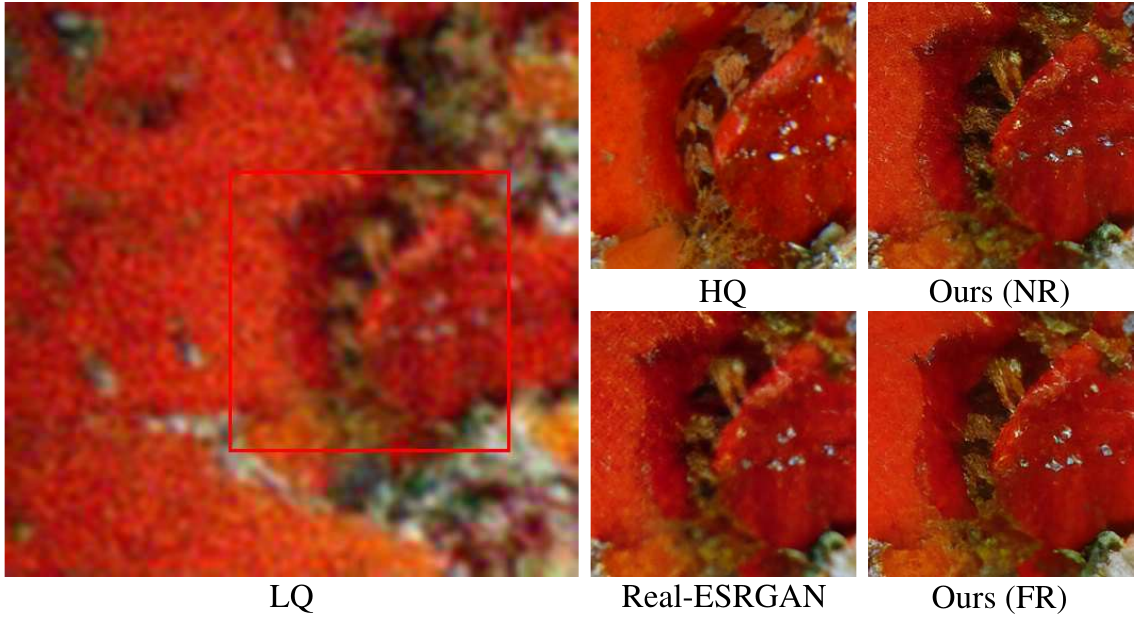}
\end{overpic}

\caption{Visual comparison of the ablation study using MDIQA as a loss function for image restoration.}
\label{fig:rest_wwo_nr_fr}

\end{center}}
\end{figure}

\subsubsection{Effect of IQA Loss}

To assess the effectiveness of our IQA loss, we incorporate it into Real-ESRGAN without any adjustment to the dimension weights. Two variants are considered: the No-Reference (NR) version and the Full-Reference (FR) version.
As shown in Table~\ref{nr_fr_loss}, incorporating our NR loss significantly improves performance over the baseline, highlighting its effectiveness in enhancing perceptual quality. 
The FR loss further boosts results, demonstrating the benefits of reference information for more accurate quality assessment and improved restoration.
As shown in the visual comparisons in Fig. \ref{fig:rest_wwo_nr_fr}, our approach yields visibly superior results.
Images restored using our loss exhibit clearer textures and improved perceptual quality compared to the baseline setting.
These visual improvements further confirm the efficacy of our method in image restoration.

\begin{table}[!htbp]
    \small
    \centering
    
    \caption{Performance comparison of NR-IQA metrics with our loss integration.}
    \label{nr_fr_loss}
    
    \scalebox{0.95}{
    \begin{tabulary}{\textwidth}{@{\hspace{2pt}}C@{\hspace{6pt}}C@{\hspace{6pt}}C@{\hspace{6pt}}C@{\hspace{6pt}}C@{\hspace{2pt}}}
        \toprule
        Loss & NIQE$\downarrow$ & CLIP-IQA$\uparrow$ & MUSIQ$\uparrow$ & Ours-Overall$\uparrow$ \\
        \midrule
        $\mathcal{L}_{org}$ & 4.2060 & 0.5841 & 62.45 & 0.5259 \\
        $\mathcal{L}_{org}+\mathcal{L}_{NR}$ & 4.0525 & 0.6586 & 66.57 & 0.6262 \\
        $\mathcal{L}_{org}+\mathcal{L}_{FR}$ & 3.7524 & 0.6991 & 68.75 & 0.6443 \\
        \bottomrule
    \end{tabulary}}

\end{table}

\subsubsection{Effect of Dimension Weight Adjustment}

To further investigate the impact of dimension weight adjustment, we conduct experiments by varying the weights of two representative dimensions, sharpness and noisiness , in both the NR and FR versions of our loss function.

\begin{table}[!htbp]
    \small
    \centering
    \caption{Performance comparison across progressively increasing sharpness ratios for both NR and FR loss variants.}
    \label{tab:sharpness_comparison}
    \renewcommand{\arraystretch}{1.0}
    
    \scalebox{0.85}{
    \begin{tabulary}{\textwidth}{@{\hspace{2pt}}C@{\hspace{6pt}}C@{\hspace{6pt}}C@{\hspace{6pt}}C@{\hspace{6pt}}C@{\hspace{6pt}}C@{\hspace{2pt}}}
        \toprule
        Type & Ratio & CLIP-IQA$\uparrow$ & MUSIQ$\uparrow$ & Ours-Sharpness$\uparrow$ & Ours-Overall$\uparrow$ \\
        \midrule
        \multirow{4}{*}{NR} & 1.0 & 0.6586 & 66.57 & 0.7265 & 0.6262 \\
                            & 2.0 & 0.7002 & 67.83 & 0.7459 & 0.6526 \\
                            & 3.0 & 0.7228 & 68.92 & 0.7719 & 0.6846 \\
                            & 4.0 & 0.7425 & 70.43 & 0.7975 & 0.7244 \\
        \midrule
        \multirow{4}{*}{FR} & 1.0 & 0.6991 & 68.75 & 0.7636 & 0.6443 \\
                            & 1.5 & 0.7035 & 68.91 & 0.7696 & 0.6498 \\
                            & 2.0 & 0.7051 & 68.52 & 0.7680 & 0.6458 \\
                            & 2.5 & 0.7098 & 69.25 & 0.7710 & 0.6503 \\
        \bottomrule
    \end{tabulary}}
\end{table}

\begin{figure}[]
\scriptsize{
\begin{center}

\begin{overpic}[width=0.45\textwidth]{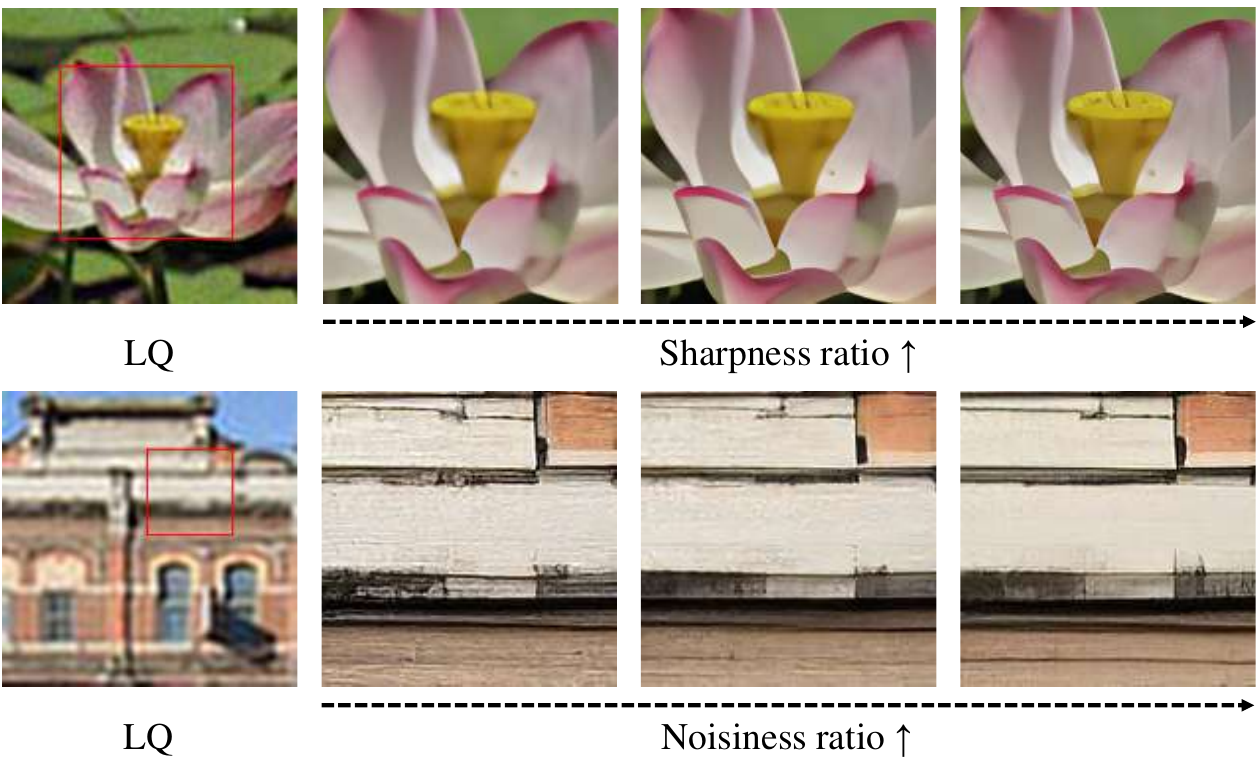}
\end{overpic}

\caption{Perceptual effects of progressively adjusting the sharpness and noisiness ratios independently in our NR loss.}
\label{fig:rest_nr_ratio}

\end{center}}
\end{figure}

\begin{figure}[]
\scriptsize{
\begin{center}

\begin{overpic}[width=0.45\textwidth]{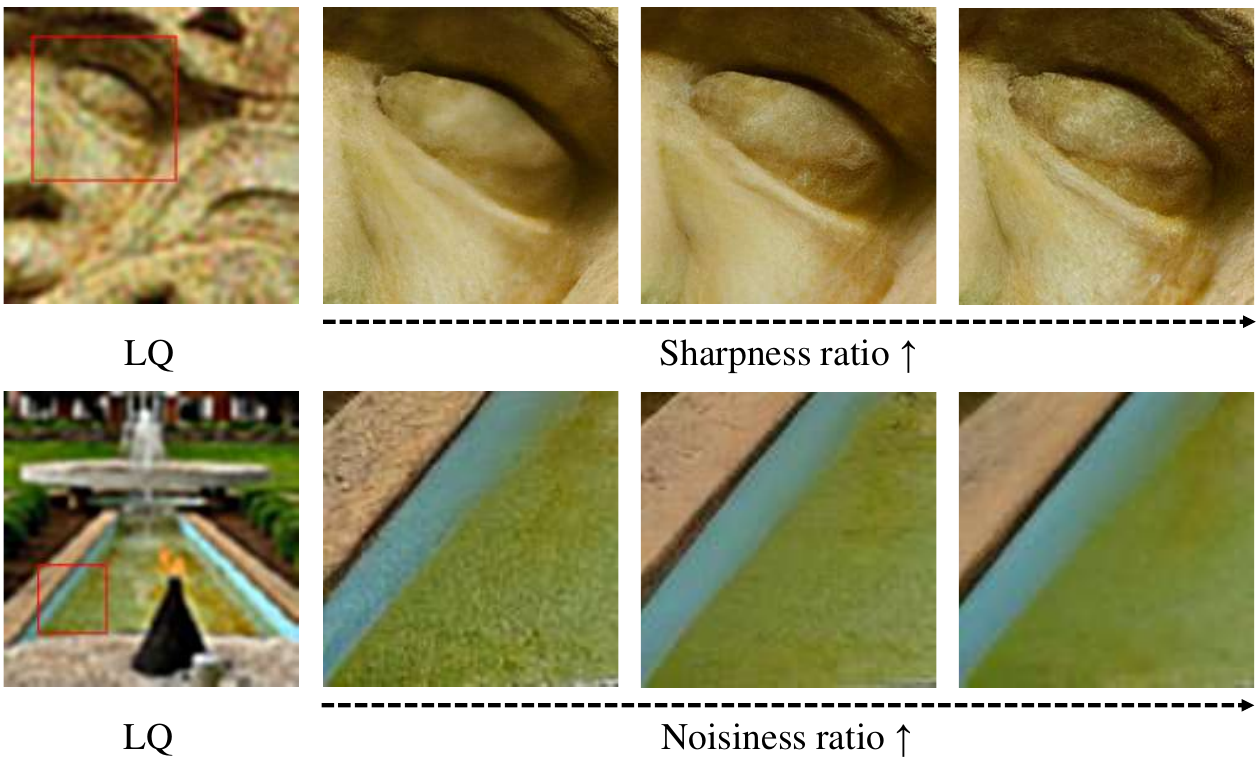}
\end{overpic}

\caption{Perceptual effects of progressively adjusting the sharpness and noisiness ratios independently in our FR loss.}
\label{fig:rest_fr_ratio}

\end{center}}
\end{figure}

For the sharpness dimension, increasing its weight ratio consistently leads to improvements across all evaluation metrics.
As shown in Table \ref{tab:sharpness_comparison}, higher sharpness weights result in better NR-IQA metrics, as well as enhanced sharpness and overall quality of MDIQA.
Visual comparisons further support these findings.
Fig. \ref{fig:rest_nr_ratio} and Fig. \ref{fig:rest_fr_ratio} confirm that increased sharpness weights produce images with progressively clearer and more detailed textures, demonstrating that prioritizing sharpness significantly improves visual appeal.

\begin{table}[!htbp]
    \small
    \centering
    \caption{Performance comparison across progressively increasing noisiness ratios for both NR and FR loss variants.}
    \label{tab:noise_comparison}
    \renewcommand{\arraystretch}{1.0}
    
    \scalebox{0.85}{
    \begin{tabulary}{\textwidth}{@{\hspace{2pt}}C@{\hspace{6pt}}C@{\hspace{6pt}}C@{\hspace{6pt}}C@{\hspace{6pt}}C@{\hspace{6pt}}C@{\hspace{2pt}}}
        \toprule
        Type & Ratio & CLIP-IQA$\uparrow$ & MUSIQ$\uparrow$ & Ours-Noisiness$\uparrow$ & Ours-Overall$\uparrow$ \\
        \midrule
        \multirow{4}{*}{NR} & 1.0 & 0.6586 & 66.57 & 0.6921 & 0.6262 \\
                            & 1.1 & 0.6562 & 66.72 & 0.6908 & 0.6160 \\
                            & 1.5 & 0.6677 & 66.51 & 0.6933 & 0.6159 \\
                            & 1.7 & 0.6625 & 66.71 & 0.6949 & 0.6156 \\
        \midrule
        \multirow{4}{*}{FR} & 1.0 & 0.6991 & 68.75 & 0.7230 & 0.6443 \\
                            & 1.1 & 0.7082 & 68.91 & 0.7236 & 0.6418 \\
                            & 1.5 & 0.7106 & 69.07 & 0.7260 & 0.6487 \\
                            & 2.0 & 0.7159 & 69.78 & 0.7303 & 0.6415 \\
        \bottomrule
    \end{tabulary}}
\end{table}

As shown in Table \ref{tab:noise_comparison}, increasing the weight ratio in the noise dimension yields only marginal gains in the overall score and other quantitative metrics.
Nevertheless, visual comparisons in Fig. \ref{fig:rest_nr_ratio} and Fig. \ref{fig:rest_fr_ratio} reveal the higher noise weights effectively reduce noise levels, producing smoother results.
This indicates that a reduction in noise levels is not consistently associated with improved perceptual quality. In certain cases, a moderate amount of noise can enhance visual realism and contribute to a more natural appearance.

    


    


\section{Conclusion}
\label{sec:conclusion}

In this work, we propose a multi-dimensional image quality assessment (MDIQA). By incorporating multiple perceptual dimensions, our approach delivers a more comprehensive, interpretable, and accurate evaluation of image quality.
We further introduce a tunable loss function for image restoration (IR), enabling dynamic adjustment of dimension weights to meet specific application needs and delivering more targeted and adaptable restoration results.
Extensive experiments demonstrate the effectiveness of our approach, achieving state-of-the-art performance on multiple IQA benchmarks and showing strong applicability in IR.
Our work highlights the potential of multi-dimensional IQA in advancing both image quality assessment and restoration tasks, offering meaningful insights for future research in perceptual image quality modeling.

\bibliography{aaai2026}

\end{document}